\documentclass[runningheads]{llncs}
\usepackage[T1]{fontenc}
\usepackage{booktabs} 
\usepackage{makecell}
\usepackage{amsmath}
\usepackage{graphicx}
\usepackage{url}
\usepackage{comment}
\usepackage[hidelinks]{hyperref}

\newsavebox\CBox

\begin{document}
\title{MedROI: Codec-Agnostic Region of Interest-Centric Compression for Medical Images}
\titlerunning{MedROI: Codec-Agnostic ROI Compression for Medical Images}
% If the paper title is too long for the running head, you can set
% an abbreviated paper title here
%
\author{Jiwon Kim\inst{1}\orcidID{0009-0000-0777-0305} \and
Ikbeom Jang\inst{1,2}\thanks{Corresponding author.}\orcidID{0000-0002-6901-983X}}
\authorrunning{J. Kim et al.}
% First names are abbreviated in the running head.
% If there are more than two authors, 'et al.' is used.
%
\institute{Princeton University, Princeton NJ 08544, USA \and
Springer Heidelberg, Tiergartenstr. 17, 69121 Heidelberg, Germany
\email{lncs@springer.com}\\
\url{http://www.springer.com/gp/computer-science/lncs} \and
ABC Institute, Rupert-Karls-University Heidelberg, Heidelberg, Germany\\
\email{\{abc,lncs\}@uni-heidelberg.de}}

\institute{Division of Computer Engineering,
Hankuk University of Foreign Studies, Yongin, Republic of Korea \and
Department of Computer Science and Engineering,
Hankuk University of Foreign Studies, Seoul, Republic of Korea \\
\email{\{harryy5981, ijang\}@hufs.ac.kr} 
}

\maketitle              % typeset the header of the contribution
% 본 연구의 강점
% 다양한 medical image processing 분야에서 image compression 연구 비교적 underexplored 이다. 우리는 이 갭을 채우겠다.
% 압축률과 압축시간 모두 개선함.
% codec-agnostic. plug-and-play

\begin{abstract}
Medical imaging archives are growing rapidly in both size and resolution, making efficient compression increasingly important for storage and data transfer. Most existing codecs compress full images/volumes (including non-diagnostic background) or apply differential ROI coding that still preserves background bits. We propose \textit{MedROI}, a codec-agnostic, plug-and-play ROI-centric framework that discards background voxels prior to compression. MedROI extracts a tight tissue bounding box via lightweight intensity-based thresholding and stores a fixed 54-byte metadata record to enable spatial restoration during decompression. The cropped ROI is then compressed using any existing 2D or 3D codec without architectural modifications or retraining.
We evaluate MedROI on 200 T1-weighted brain MRI volumes from ADNI using 6 codec configurations spanning conventional codecs (JPEG2000 2D/3D, HEIF) and neural compressors (LIC\_TCM, TCM+AuxT, BCM-Net, SirenMRI). MedROI yields statistically significant improvements in compression ratio and encoding/decoding time for most configurations (two-sided \textit{t}-test with multiple-comparison correction), while maintaining comparable reconstruction quality when measured within the ROI; HEIF is the primary exception in compression-ratio gains. For example, on JPEG2000-2D (lv3), MedROI improves CR from 20.35 to 27.37 while reducing average compression time from 1.701\,s to 1.380\,s. Code is available at \url{https://github.com/labhai/MedROI}.

\end{abstract}
%
% Medical image compression remains underexplored; however, it is becoming increasingly important as dataset size and image resolution increase. Existing compression methods either process entire images/volumes (including the backgrounds) or use differential compression that still retains background data. 
%We propose MedROI, a codec-agnostic region-of-interest (ROI) framework that focuses on the ROI containing tissue and assumes the background contains no information. We evaluate this framework on 11 existing image compression algorithms, demonstrating broad applicability across 2D and 3D image codecs. 
%Experiments on brain MRI data from 200 subjects reveal that MedROI achieves a significantly higher compression ratio (CR) and significantly decreased compression and decompression times, with minimal-to-no degradation in image quality within the ROI in most codecs. For example, CR increases from 20.4 to 27.4 while compression time reduces from 1.7 to 1.4 seconds on average when applied to JPEG2000-lv3.
%This work establishes MedROI as a practical plug-and-play solution for storing and transferring large medical image datasets. Code is available at \url{https://github.com/labhai/MedROI}.
%
%
\section{Introduction}

Medical imaging has become indispensable in modern healthcare, with data volumes experiencing rapid and continuous growth. An analysis of MICCAI conference proceedings from 2011 to 2019 shows that median dataset sizes increased by a factor of 3--10 depending on imaging modality, with annual growth rates of 21--31\% \cite{Kiryati2021}. Among these modalities, brain MRI is particularly storage-intensive due to its high-resolution 3D volumetric nature, highlighting the need for efficient compression strategies. 

This growth in medical imaging volume poses significant economic and operational challenges for clinical institutions. Compression techniques must preserve diagnostically relevant details while maintaining clinically acceptable quality \cite{Tong2025},\cite{Ig2017}. However, brain MRI volumes include large non-diagnostic background regions that contribute little to clinical interpretation but account for substantial storage overhead. 

Existing medical image compression research follows two dominant directions. Conventional full-volume compression methods apply uniform compression across the entire 3D volume using codecs such as JPEG2000 \cite{jpeg} or more recent neural compression models including TCM \cite{tcm} and BCM-Net \cite{bcm}. These methods process all voxels indiscriminately, allocating computational and storage resources to both tissue and empty background regions. 

Alternatively, Region-of-Interest (ROI) based differential compression has been explored in recent work, \cite{Li2025}, \cite{Tong2025}. These methods preserve ROIs with lossless or high-quality compression while applying aggressive lossy compression to the background. For example, wavelet-based differential techniques \cite{Rosaline2024} maintain quality in tissue regions while compressing surrounding areas at lower bitrates. Nevertheless, these methods still retain background data in compressed form, resulting in unnecessary storage usage for non-diagnostic regions. 

Both approaches, however, present opportunities for further storage optimization. Full-volume methods allocate compression resources uniformly across the entire volume regardless of clinical relevance, while differential ROI approaches still retain background data in compressed form. In clinical workflows where only tissue-containing regions are diagnostically necessary, complete removal of background voxels prior to compression offers a more storage-efficient alternative.

\begin{figure*}[!htb]
\centering
\includegraphics[width=0.87\textwidth]{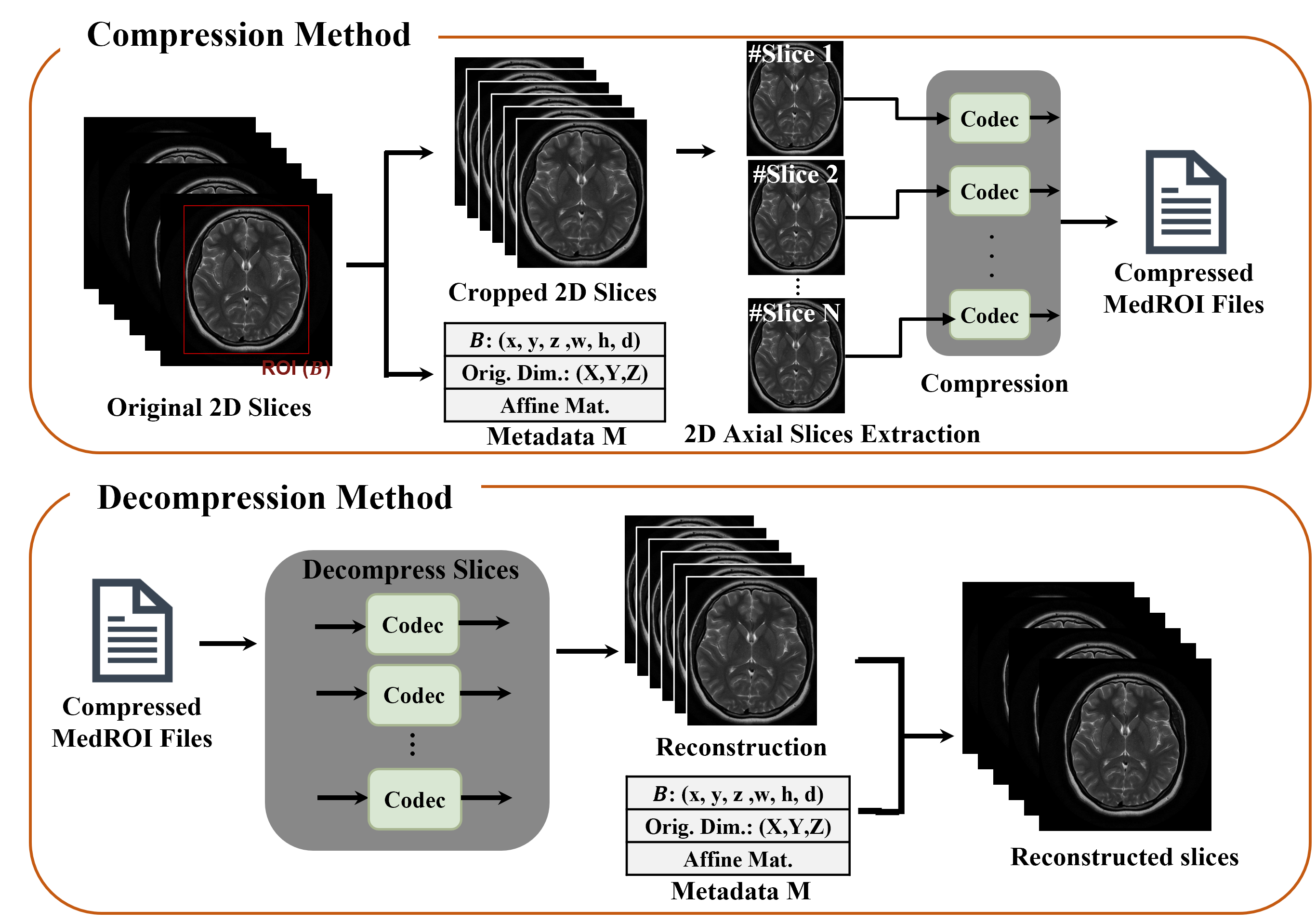}
\makebox[\linewidth][c]{\small (a) MedROI Flowchart for 2D Images}
\includegraphics[width=0.87\textwidth]{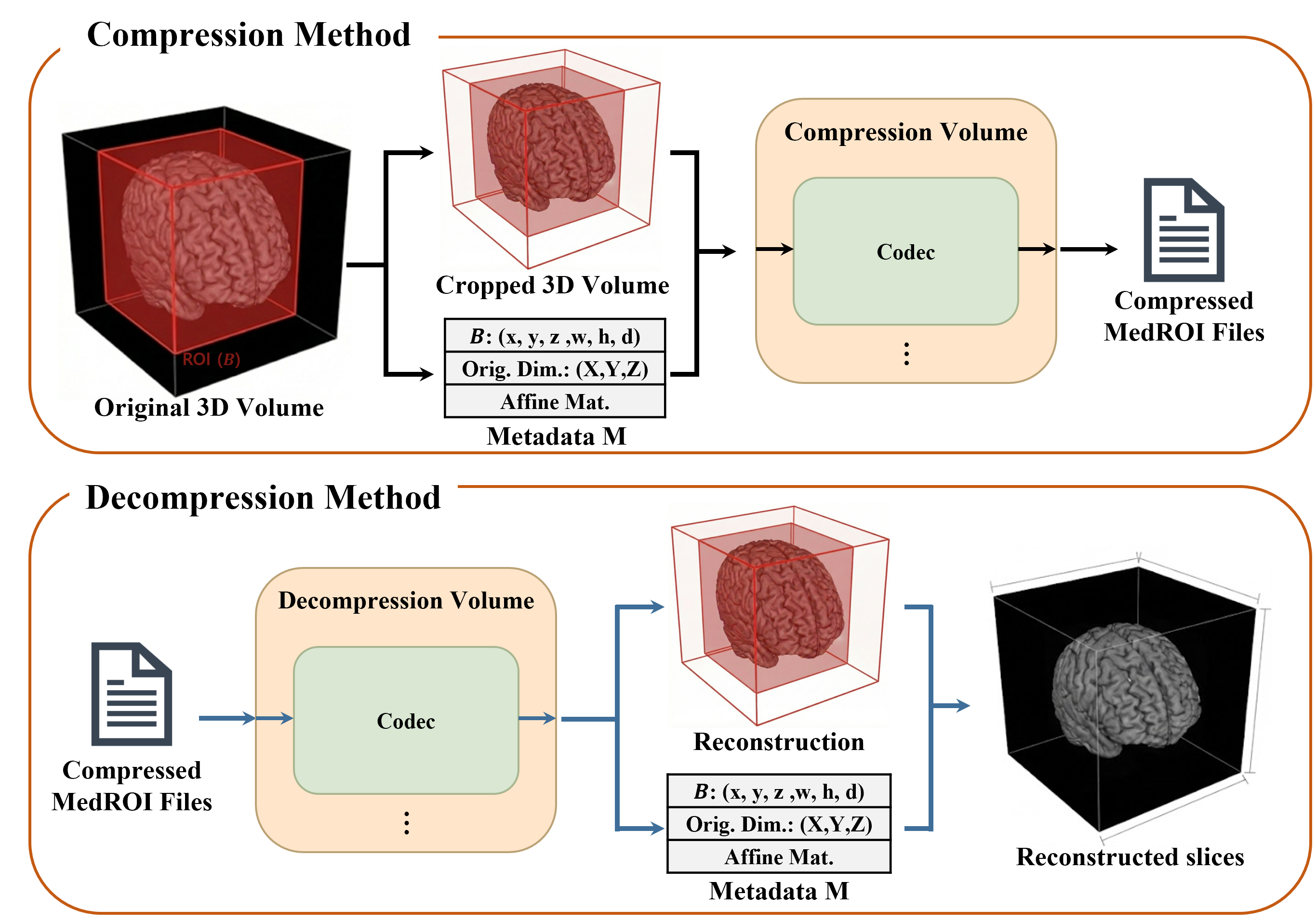}
\makebox[\linewidth][c]{\small (b) MedROI Flowchart for 3D Volume}
\caption{ROI-centric medical image compression. (a) 2D slice-wise compression pipeline: individual axial slices are extracted from the cropped volume and compressed/reconstructed independently. (b) 3D volumetric compression pipeline: ROI extraction, metadata generation, and volume-level codec compression/reconstruction. Both pipelines use metadata M (54 bytes) storing bounding box coordinates, original volume dimensions, and affine matrix for spatial restoration during decompression. }
\label{fig:method}
\end{figure*}

To address these limitations, we propose MedROI, a codec-agnostic ROI-based compression framework (fig.~\ref{fig:method}) that extracts compact bounding boxes surrounding brain tissue, effectively eliminating background regions prior to compression. The framework stores lightweight metadata (54 bytes) sufficient for spatial reconstruction and applies compression exclusively to tissue-containing voxels. Importantly, MedROI integrates seamlessly with conventional codecs, 3D volumetric methods, and 2D neural compression models without requiring architectural modifications, enabling easy deployment across diverse clinical environments. The contributions of this work are threefold: 
\vspace{-0.25cm}
\begin{itemize} \item We introduce MedROI, a generalizable ROI-based compression framework applicable to diverse compression algorithms spanning conventional codecs, 3D volumetric methods, and 2D neural networks. \item We present a systematic evaluation across six compression algorithms (JPEG2000 3D/2D, SirenMRI, HEIF, BCM-Net, TCM, TCM+AuxT) on the ADNI brain MRI dataset, showing that MedROI provides consistent compression gains across both conventional and neural codecs.\item We demonstrate that combining MedROI with the TCM neural codec achieves 81.3:1 compression ratio while maintaining clinically acceptable image quality. \end{itemize}

\section{Related Work}
\subsection{Medical Image Compression}
\paragraph{}Medical image compression methods can be categorized based on their dimensionality. 3D volumetric approaches process entire volumes to exploit inter-slice correlations. JPEG2000 3D extends the standard JPEG2000 codec to 3D wavelet transforms, enabling compression across spatial dimensions. SirenMRI [siren] leverages sinusoidal activation networks to learn continuous implicit representations of 3D MRI volumes, demonstrating effective compression while preserving structural continuity.

\paragraph{}2D compression methods process medical images slice-by-slice. Conventional codecs include JPEG2000 2D \cite{jpeg} and HEIF \cite{heif}, which are widely adopted standard-based approaches. Recent deep learning methods have achieved state-of-the-art performance, including TCM \cite{tcm}, which combines transformer-based context modeling with CNN encoders for learned image compression, TCM+AuxT \cite{tcm+auxt}, which introduces disentangled training with auxiliary transformers to improve rate-distortion performance and BCM-Net \cite{bcm}, which applies bilateral context modeling to enhance 3D medical image compression through improved 2D slice processing.

\subsection{ROI-based Compression} 
Prior work on ROI-based compression has primarily focused on differential compression strategies, where diagnostically important regions receive higher quality encoding while background regions are compressed at lower rates \cite{Ting2015}, \cite{Li2025}. JPEG2000 provides built-in ROI modes such as Maxshift and scaling-based encoding \cite{jpeg}. However, these approaches retain background information at reduced quality, limiting potential compression gains. Unlike prior methods that apply differential quality, our MedROI framework completely removes background regions, enabling a fundamentally different compression paradigm. This approach has not been systematically explored across diverse compression algorithms, particularly regarding algorithm-dependent effectiveness.

\subsection{Quality Requirements for Medical Images}
Quality assessment in medical image compression requires balancing compression efficiency with diagnostic utility. Recent studies have demonstrated that lossy compression with PSNR values ranging from 30 to 42 dB is acceptable for various medical imaging tasks \cite{Shei2025}and PSNR values around 30 dB are commonly used as a practical reference point for visually and diagnostically meaningful reconstructions \cite{Prakash2021}, \cite{Baiee2024}. Professional organizations provide varying recommendations for brain imaging: the German Roentgen Society suggests 5:1 compression for brain CT, while the Royal College of Radiologists recommends 5:1 for MRI \cite{Liu2017}. These guidelines reflect the diagnostic importance of neuroimaging, though specific PSNR thresholds vary by application. In this work, we adopt a PSNR of 30 dB or higher as our quality criterion, consistent with recent literature.

\section{Methods}

\subsection{ROI Extraction} 
\subsubsection{Preprocessing and Thresholding} We extracted brain regions from MRI volumes using an intensity-based thresholding strategy. For each volume, we first computed the mean intensity of all non-zero voxels, denoted as $\Omega_{\mathrm{nz}}$, and defined the threshold $\tau$ as:
$$\tau = \frac{1}{|\Omega_{\mathrm{nz}}|} \sum_{x \in \Omega_{\mathrm{nz}}} I(x).$$

Voxels with intensities greater than or equal to this threshold were classified as brain tissue, forming the set:
$$S = \{\, x \mid I(x) \ge \tau \,\}.$$

A 3D axis-aligned bounding box $\mathcal{B} = (x_{\min}, x_{\max}, y_{\min}, y_{\max}, z_{\min}, z_{\max})$ was then computed by taking the minimum and maximum coordinates over all elements in $S$:
$$x_{\min} = \min_{x \in S} x_x,\quad x_{\max} = \max_{x \in S} x_x,$$
$$y_{\min} = \min_{x \in S} x_y,\quad y_{\max} = \max_{x \in S} x_y,$$
$$z_{\min} = \min_{x \in S} x_z,\quad z_{\max} = \max_{x \in S} x_z.$$

To ensure complete brain coverage, we applied adaptive padding when the initial bounding box showed potential tissue loss. Specifically, if it misses more than 0.2\% of the brain tissue (miss rate > 0.002), we expanded $\mathcal{B}$ by 3 pixels in all directions:
$$\mathcal{B}_{\text{padded}} = (x_{\min}-3, x_{\max}+3, y_{\min}-3, y_{\max}+3, z_{\min}-3, z_{\max}+3).$$

This conservative strategy achieved an average miss rate of $7.92 \times 10^{-6}$
($0.0008\%$), with a maximum miss rate of $0.021$ ($2.1\%$) in the
worst case.

% *3D brain visualizations generated using Nano Banana for illustration purposes.

\subsubsection{Metadata Design} 
\label{sec:roi_metadata}
To enable reconstruction of the full-volume image from compressed ROI data, 
we design a fixed-length metadata structure $M$ with a total size of 54 bytes. 
This metadata encodes the spatial location of the ROI and the geometric 
information of the original volume, and consists of three components. 
The type, size, and role of each component are summarized in 
Table~\ref{tab:metadata}.

Since all images in the ADNI dataset have spatial dimensions smaller than 
256 voxels along each axis, spatial coordinates and original volume dimensions 
are stored as int16 to minimize storage overhead. For the affine transformation, 
only the rotation--scaling component is stored using float32 to preserve 
sufficient numerical precision. The translation component is implicitly 
recoverable from the bounding box coordinates and the original volume shape, 
thereby avoiding redundant storage.

This 54-byte metadata overhead is included in ROI mode compression 
calculations, as detailed in Section~\ref{sec:MedROI framework}.
\begin{table}[t]
\centering
\footnotesize
\caption{Metadata structure for ROI-based reconstruction.}
\label{tab:metadata}
\begin{tabular}{llll}
\toprule
\textbf{Component} & \textbf{Fields} & \textbf{Type} & \textbf{Size (bytes)} \\
\midrule
Bounding Box & 
$\left(x_{\min}, x_{\max}, y_{\min}, y_{\max}, z_{\min}, z_{\max}\right)$ &
int16 & 12 \\

Original Shape & 
$(W, H, D)$ &
int16 & 6 \\

Affine Matrix & 
$3 \times 3$ rotation--scaling submatrix &
float32 & 36 \\
\midrule
\textbf{Total} & -- & -- & \textbf{54} \\
\bottomrule
\end{tabular}
\end{table}

\subsection{Compression Methods}
\subsubsection{MedROI Framework} 
\label{sec:MedROI framework}
We propose MedROI (ROI-Centric Medical Image Compression), a selective compression framework that applies 
standard compression algorithms exclusively to anatomically relevant 
regions while completely eliminating the background. Unlike conventional ROI-based methods that compress background at reduced quality, MedROI 
discards non-tissue regions entirely, storing only the cropped brain 
volume and the 54-byte metadata structure $M$ defined in Section~ \ref{sec:roi_metadata}. The compression pipeline proceeds as follows:
\begin{enumerate}
\item Extract ROI bounding box $\mathcal{B}$ via intensity-based 
thresholding (Section 3.1.1)
\item Generate metadata $M$ encoding $\mathcal{B}$, original volume 
dimensions, and affine matrix
\item Crop the volume to $\mathcal{B}$, discarding all background voxels
\item Extract 2D axial slices from the cropped volume
\item Compress each slice independently using the target codec 
(LIC\_TCM in our baseline experiments)
\end{enumerate}

During reconstruction, the decompressed slices are placed back into a 
zero-filled volume of the original dimensions using the spatial 
information stored in $M$.

\textbf{Compression modes.} We compare two modes:

\textit{Full-volume mode:} The entire MRI volume is compressed without
ROI extraction. The compressed size and compression ratio are:

$$|\text{Compressed}_{\text{full}}| = \sum_{i=1}^{N_{\text{slices}}} |\text{LIC\_TCM}(S_i)|,$$ $$\text{CR}_{\text{full}} = \frac{|\text{Original File}|} {|\text{Compressed}_{\text{full}}|}.$$ \textit{ROI mode:} Only the cropped region $\mathcal{B}$ is compressed, including the 54-byte metadata overhead: $$|\text{Compressed}_{\text{ROI}}| = \sum_{i=1}^{N_{\text{ROI}}} |\text{LIC\_TCM}(S_i^{\text{ROI}})| + |M|,$$ $$\text{CR}_{\text{ROI}} = \frac{|\text{Original File}|} {|\text{Compressed}_{\text{ROI}}|}.$$

where $|\text{Original File}|$, $|\text{Compressed}_{\text{full}}|$, and $|\text{Compressed}_{\text{ROI}}|$ represent the number of bytes of the entropy-coded bitstreams produced by each codec in memory. The compressed size is measured from the serialized encoded representation, including all metadata overhead.

MedROI is algorithm-agnostic and applicable to any compression codec. 
We evaluate the framework across nine methods, including conventional 
codecs (JPEG2000, HEIF), 2D neural networks (LIC\_TCM, HIFIC, BCM-Net, and 3D volumetric compression (SirenMRI), with detailed 
results in Section~ \ref{results}.

\subsection{Experimental Setup}

\subsubsection{Dataset}
We evaluate MedROI on the ADNI dataset~\cite{ADNI}, consisting of 200 T1-weighted brain MRI scans in NIfTI format. Volume dimensions vary across subjects, bounded by approximately $256 \times 256 \times 256$ voxels. Since we employ pre-trained compression models without domain-specific fine-tuning, we use all 200 subjects as the test set.
For 2D compression methods, grayscale axial slices are extracted along the $z$-axis and preprocessed according to each model's requirements (e.g., normalization, channel conversion).

\subsubsection{Compression Codecs Evaluated}
We evaluate MedROI across seven compression algorithms spanning three categories: conventional codec, 2D neural compression, and 3d volumetric neural compression.
\vspace{+0.1cm}

\noindent{Conventional codecs:}
\vspace{-0.2cm}
\begin{itemize} \item JPEG2000: Wavelet-based codec with three quality levels (low/medium/high corresponding to bitrates 100/20/5). It supports both 2D and 3D image compression.
\item HEIF: HEVC-based image format with five quality levels (quality parameters: 20/40/60/80/95). We evaluated two levels: 20 and 60 (level 1 and 3)
\end{itemize} 
\vspace{-0.11cm}

\noindent{2D neural compression:} 
\vspace{-0.2cm}
\begin{itemize} 
\item BCM-Net~\cite{bcm}: CNN-based model pre-trained on TARBIT dataset 
\item TCM+AuxT~\cite{tcm+auxt}: Enhanced TCM with auxiliary transformer, using checkpoint \texttt{model\_auxt\_0035.pth.tar} 
\item LIC\_TCM~\cite{tcm}: Transformer-based codec with $\lambda=0.0035$, $N=64$, $M=320$, pre-trained on natural images 
\end{itemize} 
\vspace{-0.11cm}

\noindent{3D volumetric neural compression:} 
\vspace{-0.2cm}
\begin{itemize} 
\item SirenMRI~\cite{siren}: Implicit neural representation for 3D medical volumes 
\end{itemize}
%\vspace{-0.11cm}

\subsubsection{Implementation Details}
All methods used pre-trained checkpoints in the official codes without fine-tuning on medical data. For 2D methods, axial slices are compressed independently with arithmetic/entropy coding enabled. Experiments used Python 3.12.3 with model-specific dependencies as specified in the official repository.
Experiments were primarily conducted on an NVIDIA RTX 4090 (24GB VRAM). Only the SirenMRI codec was evaluated on an NVIDIA H200 NVL (141GB VRAM) due to its high memory requirements.

\subsubsection{Evaluation Metrics}
We measured compression efficiency, reconstruction quality, and computational cost for each compression method with and without the MedROI approach. 
Because we were not interested in the background and replaced it with zeros, image quality was measured within the ROI for MedROI. Statistical significance testing was performed on compression ratio and compression/decompression time. A two-sided \textit{t}-test was used, and multiple comparison correction was performed.

\vspace{+0.1cm}

\noindent{Compression efficiency}
\vspace{-0.2cm}
\begin{itemize} 
\item  Compression Ratio (CR): Defined in Section 3.2.2, computed as the ratio of original file size to compressed size (including 54-byte metadata overhead for ROI mode)
\item  Bits Per Pixel (BPP): Defined as total compressed bits divided by the number of voxels in the 3D volume
\end{itemize} 
\vspace{-0.11cm}

\noindent{Reconstruction quality}
\vspace{-0.2cm}
\begin{itemize} 
\item  Peak Signal-to-Noise Ratio (PSNR, dB): For 2D methods, PSNR is computed per slice and averaged across the volume. For 3D methods, it is computed on the entire volume.
\item  Structural Similarity Index (SSIM): Perceptual similarity metric on a scale of 0 to 1, averaged using the same approach as PSNR.
\end{itemize} 
\vspace{-0.11cm}

\noindent{Computational cost:}
\vspace{-0.2cm}
\begin{itemize} 
\item Compression time (s): Total time to encode all slices/volume. For the MedROI mode, this includes the time required for ROI extraction (thresholding and bounding box detection).
\item Decompression time (s): Total time to decode all slices/volume.
\end{itemize}

\label{results}
\section{Results}

MedROI achieves a significantly higher compression ratio and significantly decreased compression and decompression times ($p^*<0.05$), with minimal-to-no degradation in image quality within the ROI in most codecs, except HEIF. Compression efficiency, reconstruction quality, and computational cost for each codec with and without the MedROI approach are summarized in Table~2 and 3.

\subsection{Compression Performance}
\paragraph{2D compression methods.}
MedROI consistently improves compression efficiency across most 2D codecs while simultaneously reducing both compression and decompression times. 
Neural codecs benefit particularly strongly from the MedROI framework: TCM achieves the highest absolute compression ratio, while TCM+AuxT and BCM-Net exhibit large relative gains, indicating that background removal significantly improves the efficiency of learned image representations. 
JPEG2000 also shows clear compression gains at multiple quality levels, confirming that MedROI is effective beyond neural models. 
HEIF is the only exception, showing marginal change, likely due to its internally optimized intra-frame coding structure.

\begin{table*}[!bht]
\centering
\footnotesize
\begin{tabular}{lccccccc}
\toprule
\textbf{Model} &
\textbf{Full/ROI} &
\textbf{CR} &
\textbf{BPP} &
\textbf{PSNR} &
\textbf{SSIM} &
\textbf{Comp.(s)} &
\textbf{Decomp.(s)} \\
\midrule
JPEG2000 lv3 & Full & 20.35 & 0.395 & 33.03 & 0.852 & 1.701 & 0.350 \\
JPEG2000 lv3 & ROI  & 27.37 & 0.397 & 30.73 & 0.859 & 1.380 & 0.265 \\
\midrule
JPEG2000 lv5 & Full & 5.165 & 1.567 & 45.14 & 0.969 & 1.722 & 0.734 \\
JPEG2000 lv5 & ROI  & 6.875 & 1.582 & 41.65 & 0.964 & 1.398 & 0.565 \\
\midrule
HEIF lv1 & Full & 39.38 & 0.241 & 29.38 & 0.842 & 0.059 & 1.5e-3\\ %0.001 \\
HEIF lv1 & ROI  & 38.54 & 0.219 & 29.25 & 0.857 & 0.054 & 1.4e-3\\ %0.001 \\
\midrule
HEIF lv3 & Full & 5.907 & 1.924 & 45.74 & 0.992 & 0.112 & 2.9e-3\\ %0.002 \\
HEIF lv3 & ROI  & 5.450 & 1.650 & 44.94 & 0.990 & 0.106 & 2.7e-3\\ %0.002 \\
\midrule
TCM+AuxT & Full & 39.67 & 0.216 & 31.10 & 0.858 & 19.80 & 21.46 \\
TCM+AuxT & ROI  & 56.75 & 0.144 & 31.18 & 0.875 & 18.00 & 19.46 \\
\midrule
TCM & Full & 79.37 & 0.105 & 37.59 & 0.903 & 20.56 & 21.41 \\
TCM & ROI  & 81.31 & 0.102 & 30.89 & 0.849 & 17.84 & 18.60 \\
\midrule
BCM-Net & Full & 1.35 & 11.96 & 97.34 & 1.000 & 410.0 & 15.06 \\
BCM-Net & ROI  & 3.55 & 9.313 & 97.07 & 1.000 & 388.2 & 9.062 \\

\bottomrule
\end{tabular}
\caption{Average performance of 2D image compression methods with and without the MedROI approach. Compression and Decompression times are in seconds. comp.=compression time, decomp.=decompression time, CR=compression ratio, BPP=bits per pixel.}
\label{tab:compression_results}
\end{table*}

\label{sec:compression_performance}
\begin{table*}[!htb]
\centering
\footnotesize
\begin{tabular}{lccccccc}
\toprule
\textbf{Model} &
\textbf{Full/ROI} &
\textbf{CR} &
\textbf{BPP} &
\textbf{PSNR} &
\textbf{SSIM} &
\textbf{Comp.(s)} &
\textbf{Decomp.(s)} \\
\midrule
JPEG2000 lv1 & Full & 2.095 & 4.058 & 53.20 & 0.993 & 9.106 & 1.013 \\
JPEG2000 lv1 & ROI  & 2.316 & 3.621 & 52.38 & 0.994 & 6.729 & 0.233 \\
\midrule
JPEG2000 lv3 & Full & 4.758 & 1.755 & 47.06 & 0.981 & 5.898 & 1.023 \\
JPEG2000 lv3 & ROI  & 4.979 & 1.676 & 46.09 & 0.983 & 5.913 & 0.213 \\
\midrule
JPEG2000 lv5 & Full & 13.63 & 0.615 & 38.72 & 0.953 & 2.742 & 1.009 \\
JPEG2000 lv5 & ROI  & 14.23 & 0.591 & 37.49 & 0.949 & 2.446 & 0.204 \\
\midrule
SirenMRI & Full & 176.8 & 0.365 & 31.74 & 0.696 & 720.7 & 1.000 \\
SirenMRI & ROI  & 177.8 & 0.364 & 32.08 & 0.780 & 540.6 & 1.310 \\
\bottomrule
\end{tabular}
\caption{Average performance of 3D image compression methods evaluated with and without the MedROI approach. Total compression and decompression times are reported in seconds. comp.=compression time, decomp.=decompression time, CR=compression ratio, BPP=bits per pixel.}
\label{tab:3d_compression_results}
\end{table*}

\vspace{-0.2cm}

\paragraph{3D volumetric neural compression.} 
MedROI exhibits a significantly higher compression ratio across all 3D codecs. Also, it shows significantly reduced compression and decompression times across all codecs.
The neural implicit model SirenMRI achieves extremely high compression ratios, and MedROI further enhances its efficiency by removing non-informative background regions, leading to additional bitrate reduction and substantial speedups.
Wavelet-based JPEG2000 also benefits from MedROI at all quality levels, confirming that ROI-aware volumetric compression is effective for both neural and conventional 3D codecs.

\paragraph{}
Overall, MedROI demonstrates broad effectiveness across different codec architectures, improving compression efficiency and reducing processing times for most evaluated methods. Neural and 2D JPEG2000 methods show substantial CR improvements, while 3D methods demonstrate consistent moderate gains in both compression efficiency and computational performance.

\subsection{Rate--Distortion Analysis}
\label{sec:rd_analysis}
\begin{figure}[!bht]
    \centering
    \includegraphics[width=0.95\linewidth]{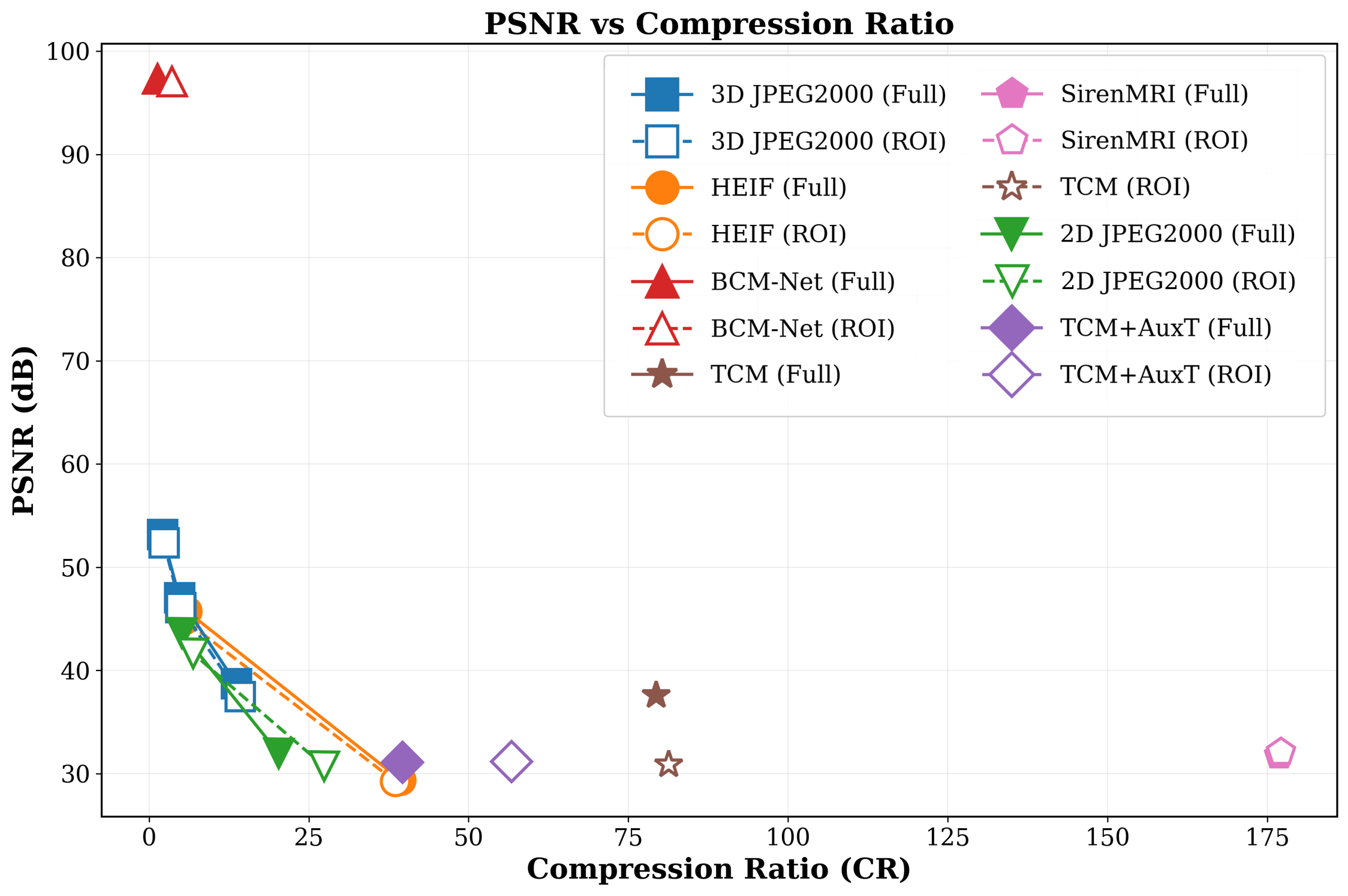}
    \caption{Rate-distortion performance of evaluated compression methods on the ADNI dataset. Each point represents the average PSNR and compression ratio across 200 brain MRI volumes. Solid markers indicate full-volume compression, while hollow markers represent ROI-based compression.}
    \label{fig:psnr_vs_cr}
\end{figure}
Figure~\ref{fig:psnr_vs_cr} illustrates the rate--distortion characteristics across all evaluated methods, demonstrating that ROI compression consistently achieves higher compression ratios than full-volume compression while maintaining comparable reconstruction quality. Across nearly all methods, the ROI curves (hollow markers) shift rightward relative to their full-volume counterparts (filled markers), indicating improved compression efficiency.

The evaluated methods occupy three distinct operating regimes in the PSNR--CR space. BCM-Net and HEIF operate in the high-quality region (PSNR $>$ 65~dB) with relatively low CR, JPEG2000 variants cover the moderate range (PSNR 30--55~dB, CR 2--30), and neural methods TCM and TCM+AuxT achieve high compression ratios (CR $>$ 40) at moderate quality levels.

The magnitude of improvement varies by algorithm architecture. Neural compression methods (TCM, TCM+AuxT, BCM-Net) show substantial rightward shifts, with 2D JPEG2000 exhibiting similar behavior. HEIF represents the only exception, where ROI and full-volume curves nearly overlap, showing minimal change in compression performance.

%\vspace{-0.2cm}

\subsubsection{Visual Quality Assessment}
\label{sec:visual_quality}
\begin{figure*}[!bht]
    \centering
    \includegraphics[width=0.75\textwidth]{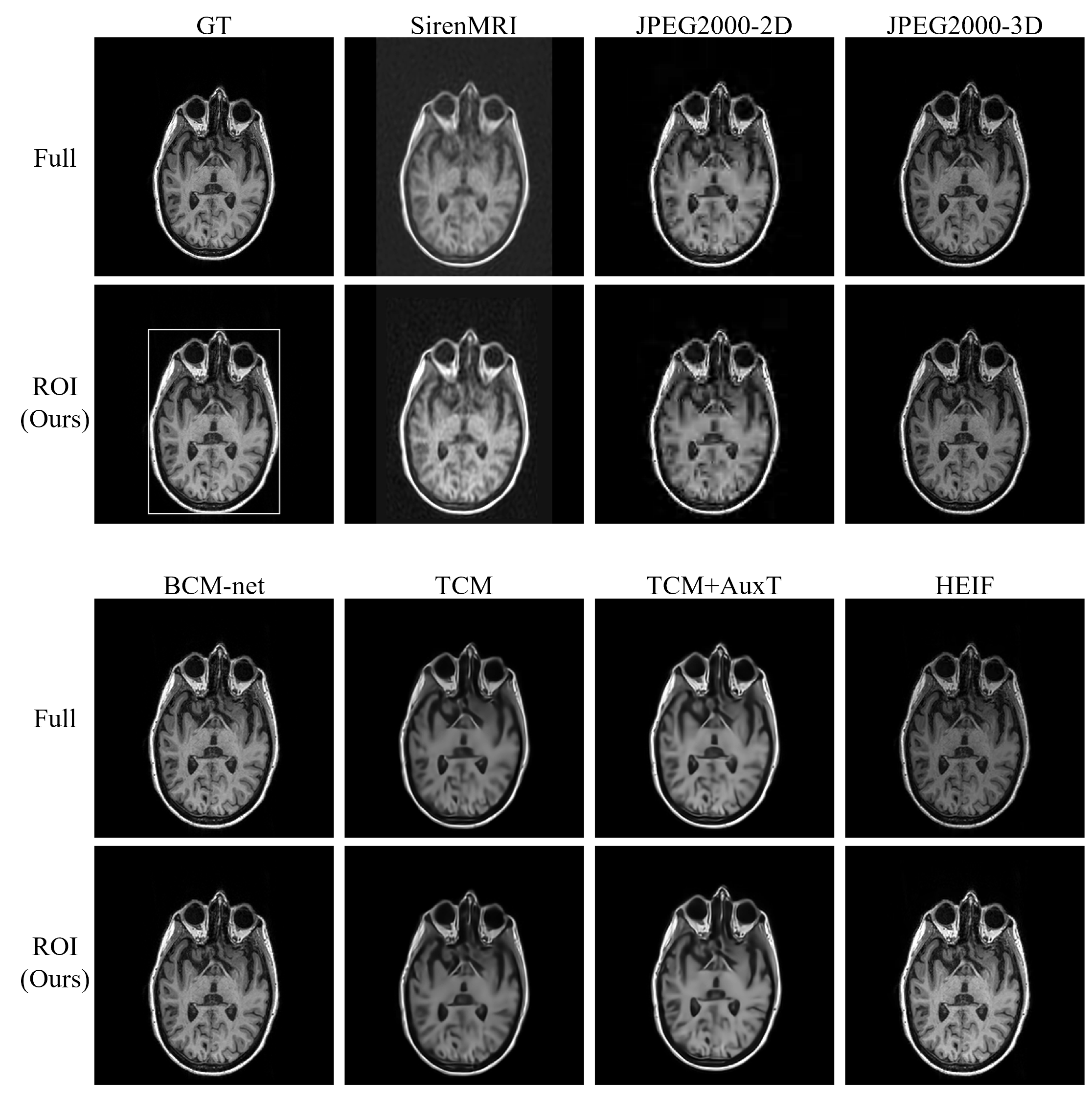}
    \caption{Qualitative comparison of Full vs ROI compression results. Visual comparison of reconstruction quality across different compression methods on subject 002\_S\_0413 (slice 128). Rows 1-2 show Full and ROI compression for 3D codecs (SirenMRI, JPEG2000-3D Lv3) and 2D conventional codec (JPEG2000-2D Lv3), while rows 3-4 display 2D neural compression codec (BCM-Net, TCM, TCM+AuxT) and HEIF Lv3. Ground truth is shown in the leftmost column, with the ROI bounding box (white rectangle) indicating the cropped region.}
    \label{fig:visual_comparison}
\end{figure*}

\vspace{-0.3cm}

The results of the sample reconstruction across different codecs are presented in Figure~\ref{fig:visual_comparison}.
3D compression methods (SirenMRI, JPEG2000-3D) produce smooth reconstructions preserving overall brain structure. 2D methods show varied characteristics: JPEG2000-2D and HEIF show high visual fidelity, while neural methods (BCM-Net, TCM, TCM+AuxT) preserve fine anatomical details. BCM-Net produces near-lossless quality closely matching the ground truth. Visual inspection shows that most algorithms produce similar reconstruction quality between full and ROI modes.

\section{Discussion \& Conclusion}
% 전체적인 결과 요약
\subsection{Compression Efficiency and Performance Gains}
The MedROI framework demonstrates substantial performance improvements across most evaluated compression algorithms. By removing non-diagnostic background regions, MedROI reduces the number of voxels requiring compression, leading to improvements in compression ratios, encoding time, and decoding time across diverse algorithm architectures.

Across evaluated methods, the framework consistently improves both compression efficiency and computational cost. Neural compression methods show substantial compression ratio gains, while conventional codecs demonstrate moderate but consistent improvements. Processing time reductions occur universally due to the decreased volume size requiring compression and decompression operations.

% 잘나온 모델&개선이 없는 모델 설명(이유)
\subsection{Algorithm-Specific Effectiveness}
\paragraph{Performance across algorithm types.}
Among 2D neural compression methods, TCM+AuxT achieves compression ratio improvement from 39.67 to 56.75 while reducing compression time from 19.80s to 18.00s and decompression time from 21.46s to 19.46s. BCM-Net demonstrates compression ratio increase from 1.35 to 3.55 with decompression time reduction from 15.06s to 9.062s. TCM maintains the highest absolute compression ratio of 81.31 (from 79.37) while reducing compression time from 20.56s to 17.84s and decompression time from 21.41s to 18.60s. Conventional 2D JPEG2000 shows consistent improvements, with level 3 achieving compression ratio increase from 20.35 to 27.37 alongside compression time reduction from 1.701s to 1.380s and decompression time from 0.350s to 0.265s.

For 3D volumetric methods, SirenMRI achieves compression time reduction from 720.7s to 540.6s while maintaining compression ratio from 176.8 to 177.8. JPEG2000 3D shows improvements across all quality levels, with level 5 increasing compression ratio from 13.63 to 14.23 while reducing compression time from 2.742s to 2.446s and decompression time from 1.009s to 0.204s.

These improvements stem from algorithm-specific characteristics. Neural networks learn data-dependent compression strategies sensitive to input distribution, and ROI cropping presents networks with concentrated anatomical content while eliminating spatially uniform background regions that consume encoding capacity. 2D JPEG2000's wavelet decomposition benefits from processing only tissue regions where transform coefficients encode meaningful anatomical variations. For 3D methods, processing smaller volumes reduces computational operations required by both volumetric wavelet decomposition and implicit neural representations.

\vspace{-0.2cm}

\paragraph{Framework versatility.}
While HEIF shows limited compression ratio improvement (level 1: 39.38$\rightarrow$38.54, level 3: 5.907$\rightarrow$5.450), it still benefits from reduced processing times. HEIF, based on HEVC/H.265, employs prediction and transform coding strategies optimized for natural images containing uniform regions. Hannuksela et al. demonstrated that HEVC achieves 97\% bit rate reduction for content where the majority of the scene remains static~\cite{Hannuksela2015}, indicating that uniform background regions require minimal bitrate. ROI cropping removes these efficiently-encoded regions, limiting compression ratio improvements. Nevertheless, the framework demonstrates effectiveness across the majority of evaluated algorithms, establishing MedROI as a versatile ROI-centric compression framework applicable to diverse compression architectures without requiring method-specific modifications.

\vspace{-0.2cm}

\subsection{Practical Applications}
The continued growth of medical imaging data creates substantial challenges for research data management and distribution. Large-scale neuroimaging studies generate terabyte-scale datasets that must be stored, maintained, and shared across research institutions. Moreover, improved comprassion ratios substntially reduce data tranfer overhead, enabling faster and more efficient sharing of large imaging datasets in multi-center collaborations and public research repositories worldwide. 

MedROI addresses these challenges through high compression efficiency and reduced processing times. For example, combining the framework with TCM on the ADNI dataset (200 volumes, 4.4GB) achieves 81.3:1 compression ratio while reducing compression time from 20.56s to 17.84s and decompression time from 21.41s to 18.60s per volume. Such compression efficiency and computational performance improvements directly translate to reduced storage infrastructure costs and faster data processing workflows.

% mammo 실험내용
We conducted a small pilot using an additional dataset, INbreast (a different modality and anatomy 
\cite{mammo}, to assess whether MedROI effects persist beyond ADNI. Mammogram data from 15 participants were evaluated using the TCM and JPEG2000 2D codecs. TCM exhibited a 75.7\% increase in CR, JPEG2000 lv1 and lv3 a 74.6\% increase, and JPEG2000 lv5 a 44.9\% increase. Compression and decompression times decreased significantly in all cases.

The framework's codec-agnostic design enables deployment across diverse institutional environments with varying computational resources and existing compression infrastructure. Institutions can select compression algorithms matching their specific requirements--whether prioritizing maximum compression efficiency with neural methods or computational simplicity with conventional codecs--while consistently benefiting from the MedROI framework.

% 한계점 : 다른 데이터셋 실험 X, ROI method, 임상적 증명필요 
\subsection{Limitations \& Future Work}
This study has several limitations that should be acknowledged. First, we evaluate MedROI primarily on the ADNI brain MRI dataset consisting of T1-weighted scans. Only a small pilot study using additional mammography data was conducted to assess generalizability. An extensive evaluation on diverse organs and modalities remains necessary to establish broader applicability.

Also, our quality assessment relies on PSNR and SSIM metrics. While PSNR > 30 dB serves as an established quality threshold for medical imaging, these metrics do not directly measure diagnostic utility. Clinical validation studies involving diagnostic accuracy assessment would provide more direct evidence of the framework's suitability for clinical deployment.

\subsection{Conclusion}
\label{sec:discussion_conclusion}
We presented MedROI, a codec-agnostic ROI-centric framework for medical image compression that removes non-diagnostic background prior to encoding and stores only a fixed 54-byte metadata record for spatial restoration. MedROI was evaluated across 6 codec configurations spanning conventional codecs and neural compressors. Across most configurations, it yields statistically significant gains in compression ratio and consistently reduces both compression and decompression time, while maintaining comparable reconstruction quality. Overall, MedROI offers a practical plug-and-play strategy for storing and transferring large medical imaging datasets without modifying existing codecs. Future work will extend validation to additional anatomies and modalities, investigate more robust ROI extraction beyond global thresholding, and incorporate task- or clinician-centered evaluation of diagnostic fidelity.

\end{document}